\documentclass[10pt,twocolumn,letterpaper]{article}
\usepackage{cvpr}
\usepackage{times}
\usepackage{epsfig}
\usepackage{graphicx}
\usepackage{float}
\usepackage{amsmath}
\usepackage{amssymb}
\usepackage{lipsum}
\usepackage{subcaption}
\usepackage[pagebackref=true,breaklinks=true,letterpaper=true,colorlinks,bookmarks=false]{hyperref}
\cvprfinalcopy 

\ifcvprfinal\fi

\begin{document}
\title{Learning to see people like people}
\author{Amanda Song\\
University of California, San Diego\\
9500 Gilman Dr, La Jolla, CA 92093\\
{\tt\small feijuejuanling@gmail.com}
\and
Linjie Li\\
Purdue University\\
610 Purdue Mall, West Lafayette, IN 47907\\
{\tt\small li2477@purdue.edu}
\and
Chad Atalla\\
University of California, San Diego\\
9500 Gilman Dr, La Jolla, CA 92093\\
{\tt\small catalla@ucsd.edu}
\and
Garrison Cottrell\\
University of California, San Diego\\
9500 Gilman Dr, La Jolla, CA 92093\\
{\tt\small gary@ucsd.edu}
}
\maketitle
\begin{abstract}
The human perceptual system can make complex inferences on faces, ranging from the objective evaluations regarding gender, ethnicity, expression, age, identity, etc. to subjective judgments on facial attractiveness, trustworthiness, sociability, friendliness, etc. Whereas the objective aspects have been extensively studied, less attention has been paid to modeling the subjective perception of faces. Here, we adapt 6 state-of-the-art neural networks pretrained on various image tasks (object classification, face identification, face localization) to predict human ratings on 40 social judgments of faces in the 10k US Adult Face Database. Supervised ridge regression on PCA of the conv5\textunderscore2 layer in VGG-16 network gives best predictions on the average human ratings. Human group agreement was evaluated by repeatedly randomly splitting the raters into two halves for each face, and calculating the Pearson correlation between the two sets of averaged ratings. Due to this methodology, the model’s correlations with the average human ratings can exceed this score. We find that 1) model performance grows as the consensus on a face trait increases, and 2) model correlations are always higher than human correlations with each other. These results illustrate the learnability of the subjective perception of faces, especially when there is consensus, and the striking versatility and transferability of representations learned for object recognition. This work has strong applications to social robotics, allowing robots to infer human judgments of each other.
\end{abstract}

\section{Introduction}
Recent advances in deep convolutional networks have driven tremendous progress in a variety of challenging face processing tasks including face recognition\cite{taigman2014deepface}, face alignment\cite{zhu2015face}, and face detection\cite{Stewart_2016_CVPR}. However, humans not only read objective properties from a face, such as gender, expression, race, age and identity, but also form subjective impressions of the social aspects of a face\cite{todorov2013social,todorov2015social}, such as facial attractiveness\cite{thornhill1999facial}, friendliness, trustworthiness\cite{todorov2008evaluating}, sociability, dominance\cite{mignault2003many}, and typicality. Despite the relative less attention received by the social perception of faces, social judgment is an important part of people's daily interactions, and it has significant impact on social outcomes, ranging from electoral success to sentencing decisions\cite{oosterhof2008functional, willis2006first}. Whereas current computer vision techniques exceed human abilities at recognizing a face and identifying the objective properties of a face \cite{taigman2014deepface, Stewart_2016_CVPR}, awareness of human subjective judgments is important for social robotics theory-of-mind inferences. Accurate predictions of social aspects of faces can help robots better understand how humans interact with and perceive each other, and can make a robot aware of inherent human biases, as these judgments rarely correspond to reality (except, perhaps,  attractiveness) \cite{todorov2015social}.  

In this paper, we teach a machine to infer social impressions, that match human judgments, from faces. We examine a list of 20 pairs of social features that are typically studied by social psychologists, and that are relevant to social interactions between people
~\cite{todorov2008understanding,todorov2015social, oosterhof2008functional}. Examples are  attractiveness~\cite{grammer1994human, thornhill1999facial, eisenthal2006facial, kagian2008machine, gray2010predicting}, trustworthiness~\cite{falvello2015robustness, todorov2008evaluating}, sociability, aggressiveness~\cite{mignault2003many}, friendliness, kindness, happiness, familiarity~\cite{peskin2004familiarity}, and memorability~\cite{bainbridge2013intrinsic, khosla2013modifying}). Although social perceptions of faces are subjective, there is often a  consensus among human raters in how they perceive facial attractiveness,  trustworthiness and dominance\cite{falvello2015robustness, eisenthal2006facial}. This indicates that faces contain high-level visual cues for social interactions, and therefore it is possible to model this process with machine learning techniques. We take advantage of the state-of-the-art neural network models trained for object recognition and face recognition tasks and use their internal representations for social perception learning. In all 40 social dimensions, our model correlates with human averaged ratings better than the humans correlate with each other. 

The contributions of the paper are summarized below:
\begin{itemize}
\item To the best of our knowledge, this work is the first attempt to systematically examine the consistency of human social perceptions of faces, to explore the landscape of social feature semantic space, and to predict human judgments of 40 social attributes of faces; 
\item We adapted 6 state of the art neural network algorithms trained for various visual tasks to make social judgment predictions on faces and achieve high correlations with human ratings in all 40 dimensions;
\item We evaluate the tuning properties of nodes in the best network and visualize the patterns that maximally ignite the perceptions for each specific social dimensions to facilitate a better understanding of the neural networks' behavior in face processing. 
\end{itemize}

The rest of the paper is organized as follows. In Section \ref{related_work}, we review related work on social perception modeling. Section \ref{method} and \ref{experiment} summarize the methodology and the experimental framework. The experimental results and visualizations are presented in Section \ref{result} and \ref{viz}. Section \ref{conclude} concludes the paper.

\section{Related work} \label{related_work}
The focus of our paper is to infer as much social judgment information as possible from a face image and to predict the subjective impression of faces by learning from human group data. We review related work in terms of the visual features they use, the dataset they choose, the evaluation metric they adopt, and the social attributes they examine. 

\textbf{Visual features} 
Since the early 1990s, psychologists have identified that high level visual features, such as the averageness of a face\cite{langlois1990attractive, rhodes2003fitting} and the symmetry of the face \cite{scheib1999facial} can explain why certain faces look more attractive.

Machine learning researchers have developed various computer vision features and models to predict social perceptions of faces, especially facial attractiveness. Yael et al.\cite{eisenthal2006facial} used geometric ratios and distances between facial features based on facial landmarks to build an attractiveness predictor (0.65 correlation with human raters, face database size=184). End-to-end neural networks were applied to predict facial attractiveness in 2010\cite{gray2010predicting} (correlation 0.458, face database size=2056, young female faces only). Amit Kagian and his colleagues have used a combination of landmark-derived features along with global features to obtain a high correlation with human group averages on facial attractiveness \cite{kagian2008machine}(0.82 Pearson correlation, face database size=91). Traditional computer vision features such as SIFT, HoG, Gabor filters have been blended to predict the relative ranking of facial attractiveness in \cite{altwaijry2013relative}(rank order correlation 0.63, face database size=200). Rothe et al. incorporate collaborative filtering techniques with visual features extracted from pretrained VGG networks\cite{simonyan2014very} to achieve individual-level prediction of facial attractiveness\cite{rothe2015some}(correlation 0.671 on female face queries, database size = 13,000). McCurrie et al. \cite{mccurrie2016predicting} build a model based on a pretrained VGG network to predict trustworthiness, dominance and IQ in faces ($R^2$ values on trustworthiness, dominance and IQ are 0.5687, 0.4601, 0.3548 respectively, face database size=6000).  Previous papers have achieved correlations with human performance between $0.458$ to $0.82$ in attractiveness predictions, depending on the dataset and method used. However, to date, there is no standard dataset that has been used to compare these approaches. 

\textbf{Dataset}
Earlier studies employ datasets with relatively small numbers of faces (a few hundred) and most face datasets use young Caucasian faces only, as pointed out by \cite{laurentini2014computer}. In contrast, the MIT dataset\cite{bainbridge2013intrinsic} we use contains 2,222 high quality color images that vary in ethnicity, gender, age and expression, with ratings on 40 attributes. This dataset is smaller than two of the ones mentioned above. The first is collected from howhot.io, an online dating website\cite{rothe2015some} and contains 13,000 face images, but that work focused on personalized prediction of facial attractiveness, rather than average ratings. There are only binary choices (like or dislike) indicating implicit preference of facial attractiveness. The second one is collected from testmybrain.com, contains 6,000 grayscale face images
~\cite{mccurrie2016predicting}, and includes just three social features: dominance, IQ and trustworthiness.

\textbf{Evaluation metric}
Social perceptions of faces are collected from human participants in various ways. The most common way is to ask for a discrete rating, say from 1-9
\cite{bainbridge2013intrinsic}, or 
1-7
\cite{eisenthal2006facial} from a number of raters, and then use the group average as the score for a face in the specified feature dimension (e.g. attractiveness). The consistency of ratings between humans is checked by repeatedly randomly splitting human participants into two subgroups and then computing the correlation between the two groups' mean ratings. To compare model predictions with human ratings, Pearson correlation\cite{kagian2008machine, dantcheva2011female}, Spearman rank correlation\cite{bainbridge2013intrinsic} and R-squared values\cite{mccurrie2016predicting} are used, 
depending on the nature of the data. Another method is to present a pair of faces or multiple faces and ask for a relative ranking in a particular dimension (e.g., attractiveness). Prediction accuracy is measured using Kendall's Tau and the Gamma Test
~\cite{altwaijry2013relative}. In Rothe et al.~\cite{rothe2015some}, a person indicates his/ her preference by choosing to  like or dislike another user's face photo.  In this paper, since our goal is to predict a continuous score of human average ratings, and our raters do not all rate the same faces, we also use Pearson correlation with average human ratings on a per-face basis.  

\textbf{Social attributes}
Although social perceptions are a subjective judgment, and may not reflect a person's actual traits or mental states, humans tend to share consensus on their first impressions. Kiapour et al.\cite{kiapour2014hipster} and Wang et al.\cite{wang2015bikers} find that the social styles of people (bikers vs. hipsters, for example) can be identified and classified from image features. Dhar et al. (2011) show that the interestingness of an image can be quantified and predicted~\cite{dhar2011high}. Bainbridge et al. (2013) prove that the memorability of a face image can be predicted and modified to make it more memorable
~\cite{bainbridge2013intrinsic}. Todorov et al. ~\cite{todorov2013validation, todorov2013social,todorov2015social} used synthesized faces to study the perception of competence, dominance, extroversion, likeability, threat, trustworthiness and attractiveness in faces~\cite{todorov2008evaluating}. However their face photos lack realism compared to real-world photos and therefore cannot predict human's social perceptions of real faces in a more natural environment. McCurrie et al. \cite{mccurrie2016predicting} have worked toward removing this limitation by using real human faces to make predictions of trustworthiness and dominance ratings. 
12
From the literature, we can observe two trends: (i) Besides McCurrie et al. ~\cite{mccurrie2016predicting} and Todorov. et al\cite{todorov2008evaluating}, most machine learning work on social perception of faces focuses on attractiveness prediction, leaving the prediction of other social perceptions largely unstudied. We aim to bridge this gap in our paper. (ii) As summarized by Laurentini et al\cite{laurentini2014computer} usually small datasets are used, with few variations on expression, gender, ethnicity and age. The dataset we chose overcomes the above limitations and has comprehensive coverage of a list of 40 social feature ratings. 

The papers closest to ours are McCurrie et al. \cite{mccurrie2016predicting} and Todorov. et al\cite{todorov2008evaluating}. Our paper differs from theirs in three major ways: (1) Todorov. et al's work is on synthesized faces whereas ours is on realistic photos; (2) McCurrie et al. \cite{mccurrie2016predicting} predict three social features, dominance, trustworthiness, and IQ, whereas we look at 40 social features including trustworthiness, aggressiveness (a term close to their dominance), and intelligence (close to their IQ term), so our feature set can be considered to be a superset of theirs; and (3) we compared various feature extraction methods, including traditional geometric features and 6 neural networks pretrained for various tasks (face identification, face localization, object recognition). We also examine the effect of fine-tuning the network compared with directly applying ridge regression on extracted features from higher layer of the networks. 

\section{Method} \label{method}
In this section we first describe the dataset used in our experiments. Next, we introduce our method for predicting social perceptions of faces. Finally, we explain how we visualize the features that contribute most to social trait predictions.

\subsection{Dataset}
To predict how human evaluate social traits of a face at a glance, we use the dataset collected by Aude Oliva's group \cite{bainbridge2013intrinsic}. The dataset consists of 2,222 images of faces sampled from the 10k US Adult Face Database and annotated for 20 pairs of social attributes. Each attribute is rated on a scale of 1-9 (1 means not at all, 9 means extremely) and each image is rated by 15 subjects. We take the average rating across all raters as a collective estimation of the social feature for every face. 

The 20 pairs of social traits are: (attractive, unattractive), (happy, unhappy), (friendly, unfriendly), (sociable, introverted), (kind, mean), (caring, cold), (calm, aggressive), (trustworthy, untrustworthy), (responsible, irresponsible), (confident, uncertain), (humble, egotistical),(emotionally stable, emotionally unstable), (normal, weird), (intelligent, unintelligent), (interesting, boring), (emotional, unemotional), (memorable, forgettable), (typical, atypical), (familiar, unfamiliar) and (common, uncommon).

\begin{figure*}[th]
\begin{subfigure}{.5\textwidth}
      \centering     
      \includegraphics[width=\linewidth]{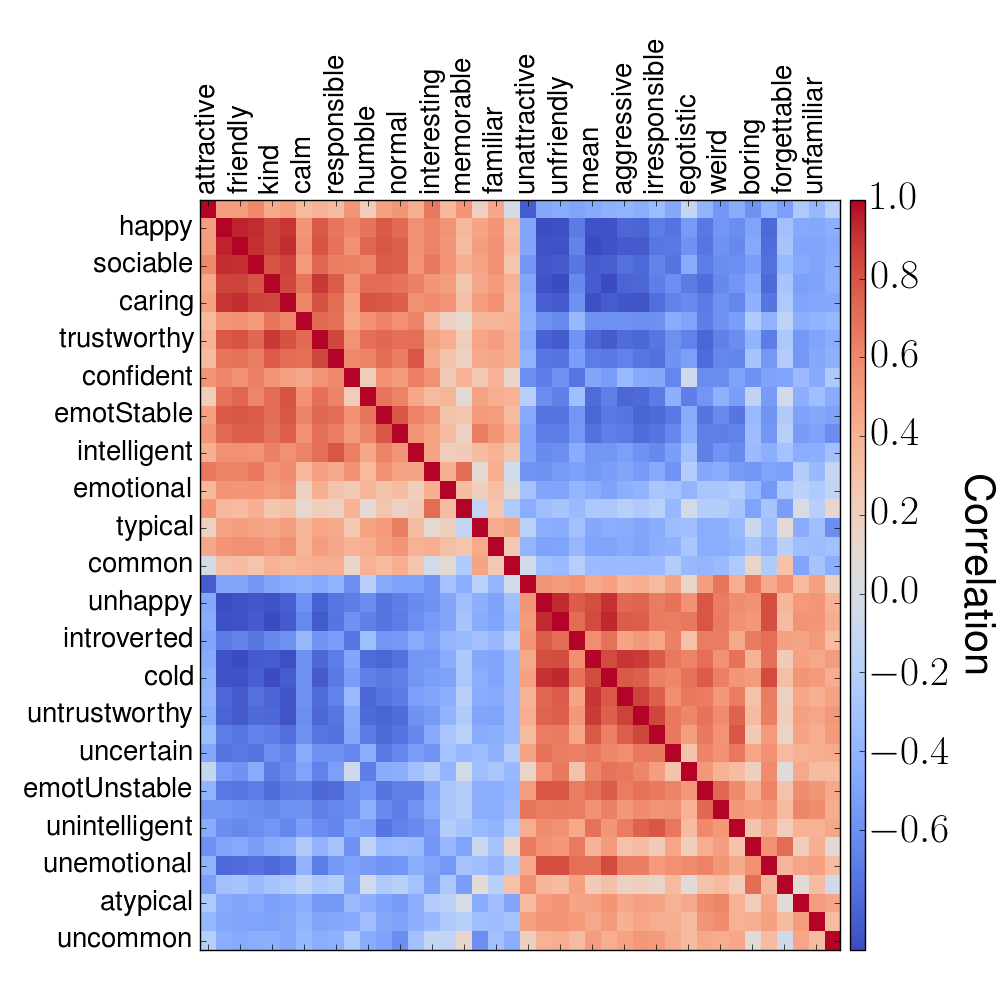}   
\end{subfigure}%
\begin{subfigure}{.5\textwidth}
      \centering     
      \includegraphics[width=\linewidth]{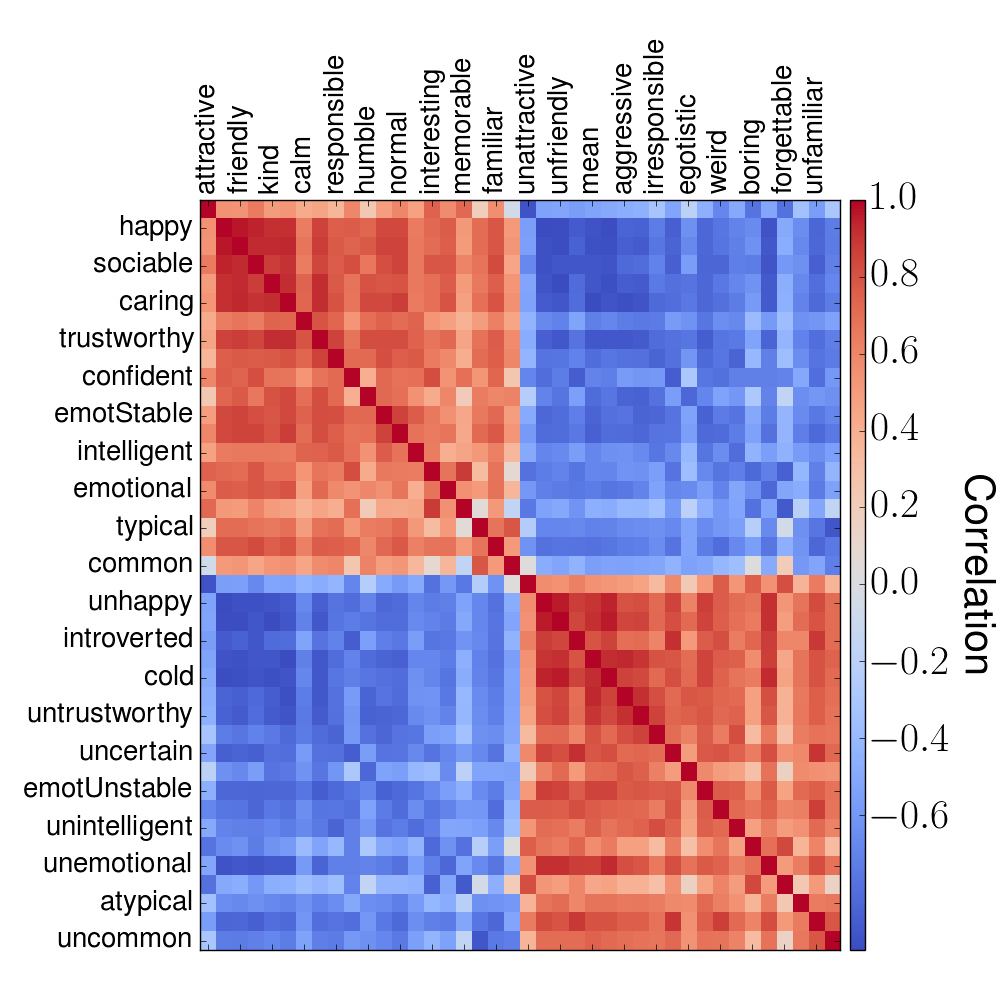}
\end{subfigure}
\caption{Correlation heatmaps among social features. Left: human. Right: network.}
\label{socialCorr}
\end{figure*}

Clearly, some of these traits will be highly correlated, and are predictable from the others. We compute the Spearman's rank correlation between every pair of social features and show their correlations in a heatmap (see the left figure in Figure \ref{socialCorr}). We put features together in the map based on similarity and positive/negativeness. From the figure, we can see that negative social features such as untrustworthy, aggressiveness, cold, introverted, irresponsible form a correlated block, while most positive features such as attractive, sociable, caring, friendly, happy, intelligent, interesting, confident are highly correlated with each other. Although we chose 20 pairs of opposite features, they are not completely complementary and redundant. Principal component analysis shows that it takes 24 principal components to cover $95\%$ of the variance. 

\subsection{Regression Model for Social Attributes}
After we average human ratings on each face, each face receives a continuous score from 1 to 9 in all social dimensions. We model these social scores with a regression model. Our proposed algorithm is a ridge regression model on features extracted from deep convolutional neural networks (CNN). Since CNN features are usually high-dimensional, we first perform Principal Component Analysis (PCA) on the extracted features of the training set to reduce the dimensionality. The PCA dimensionality is chosen by cross-validation on a validation set, separately for each trait. The PCA weights are saved and further used in fine-tuning our CNN-regression model. 

\subsection{Feature Visualization}
\label{sec:fv}
Attempting to understand what features are most helpful in social attribute prediction, we visualize the features extracted from the CNN. Two different methods were proposed in past feature visualization studies: dataset-centric methods~\cite{yosinski2015understanding, zeiler2014visualizing}, 
and a network-centric method\cite{yosinski2015understanding,yu2016deep}. 

The dataset-centric method we employed is to display image patches from the training set that cause high activation for the feature units and use the deconvolution method to highlight the portions of the image that are responsible for firing the important feature neurons ~\cite{yosinski2015understanding,zeiler2014visualizing}. 

The network-centric approach is usually used in classification networks. This method produces an image that is based on adapting the input by maximizing the output category activation using gradient ascent, i.e., it is mainly a function of the network.
 \cite{yosinski2015understanding, yu2016deep}. The key idea is to optimize the input image so that the target neuron can be highly activated. We apply this idea to the output (regression) neuron as well as the top nine neurons that influence that output individually.

\section{Experimental framework} \label{experiment}
In this section, we report our experimental framework using 6 CNN-based regression models with respect to two baselines, human correlation between groups of raters, and a baseline model using the geometric features. 

\subsection{Baseline I: Human Correlations} \label{baseline1}
Since these social attributes are all subjective perceptions rated by people, it is informative to examine to what extent people agree with each other upon those social judgments. We performed the following procedure 50 times for each attribute and then averaged the results:
\begin{enumerate}
\item For each face, we randomly split the 15 raters evenly into two groups of 7 and 8. (Note: the raters for each face will, in general, be different sets).
\item We calculate the two group's average ratings for each face, obtaining two vectors of length 2,222 (there are 2,222 faces in the dataset). 
\item We calculate the correlation between the two vectors. 
\end{enumerate}
The results are shown in the second column of Table \ref{tb:correlations}. For every social attribute, the averaged correlation between human subgroups serves as an index of the rating consistency.

\subsection{Baseline II: Regression on geometric features}
\begin{figure}[!htbp]
\centering
\begin{subfigure}{.25\textwidth}
  \centering
  \includegraphics[width=.5\linewidth]{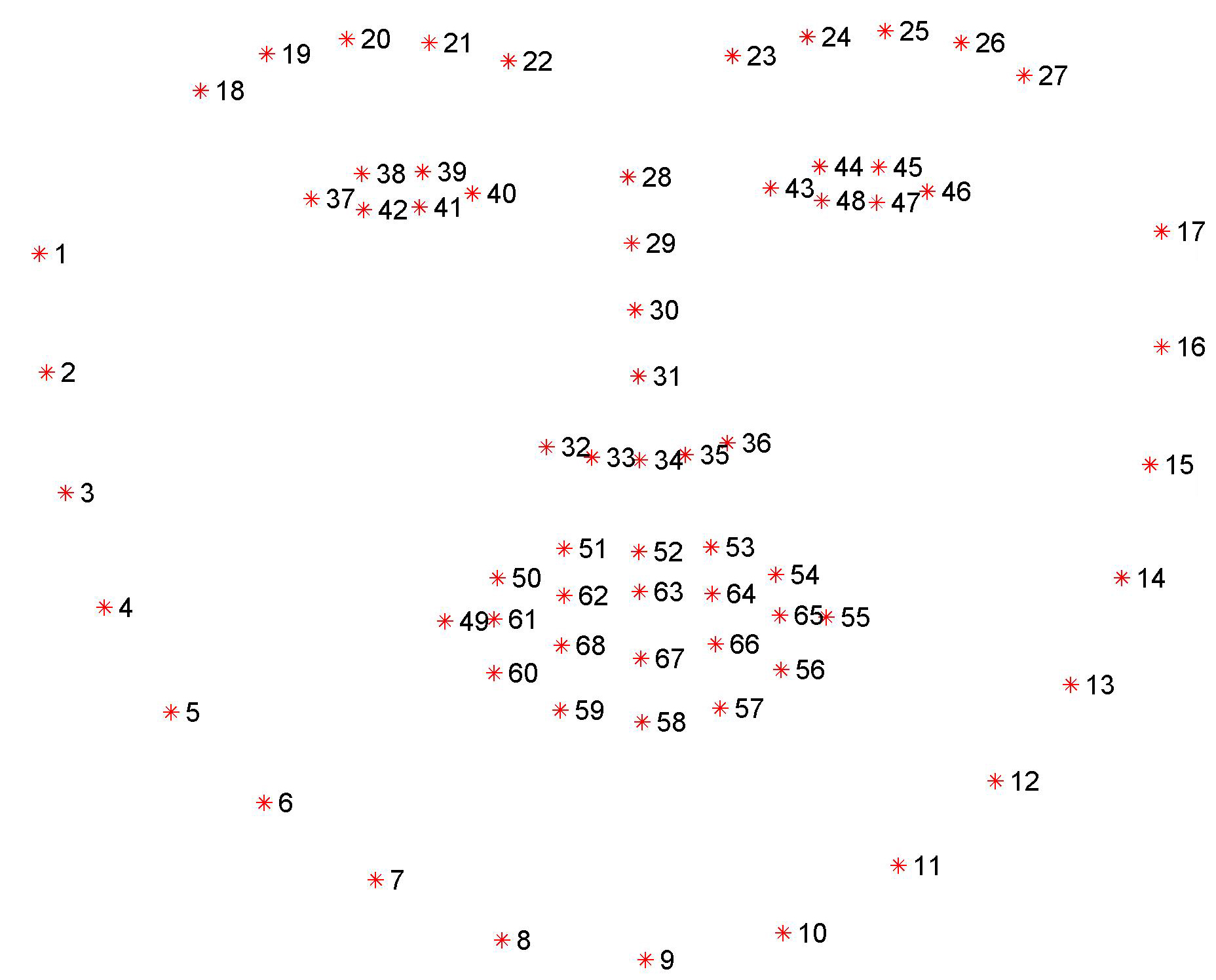}
\end{subfigure}%
\begin{subfigure}{.25\textwidth}
  \centering
  \includegraphics[width=.5\linewidth]{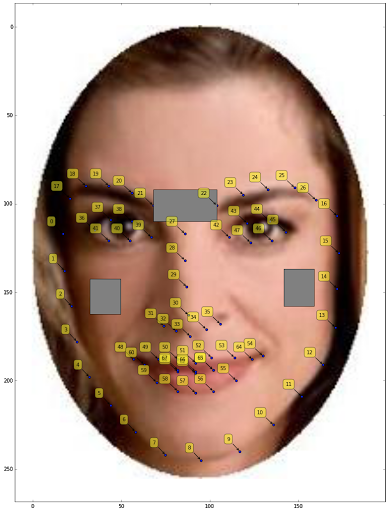}
\end{subfigure}
\caption{68 face landmarks labeled by dlib software automatically. The gray regions are the locations used for computing smoothness and skin color.}
\label{faceLandmarks}
\end{figure}

Past studies on facial attractiveness have found that attractiveness can be inferred from the geometric ratios and configurations of a face\cite{eisenthal2006facial, kagian2008machine}. We suggest that other social attributes can also be inferred from geometric features. We compute 29 geometric features based on definitions described in \cite{ma2015chicago} and further extract a "smoothness" feature and skin color features according to the procedure in \cite{eisenthal2006facial, kagian2008machine}. The "smoothness" of a face was evaluated by applying a Canny edge detector to windows from the cheek/forehead area \cite{eisenthal2006facial}. The more edges detected by edge detectors within the window, the less smooth the skin is. The regions we chose to compute smoothness and skin color are highlighted in the right subplot of Figure \ref{faceLandmarks}). The "skin color" feature is extracted from the same window as "smoothness", converted from RGB to HSV. Regressing on these handcrafted features alone are not enough to capture the richness of geometric details about a face, we therefore use a computer vision library (dlib, C++) to automatically label 68 face landmarks (see Figure \ref{faceLandmarks}) for each face and compute distances and slopes between any two landmarks. Combining 29 handcrafted geometric features, smoothness, color and the distance-slope features, we obtain 4592 features in total. Since the features are highly correlated, we apply PCA to reduce dimensionality. Again, the PCA dimensionality is chosen by cross-validating on the hold out set separately for each facial attribute. Then a ridge regression model is applied to predict social attribute ratings of a face. The hyperparameter of ridge regression is selected by leave-one-out validation within the training set. 

\subsection{CNN-based Regression Model}
We initially compared six neural network architectures: (1) VGG16, (2) VGG-Face from the Oxford Visual Geometry Group’s VGG networks\cite{simonyan2014very}, (3) AlexNet (the publicly available CaffeNet reference model) \cite{krizhevsky2012imagenet} (4) Inception from Google \cite{szegedy2015going} (5) a shallow face identification Siamese neural network that we trained from scratch: Face-SNN and (6) a state of the art VGG-derived network trained for the face landmark localization task: Face-LandmarkNN.  These comparisons were performed with the Caffe deep learning framework \cite{jia2014caffe}.

To find the best CNN to predict social attribute ratings among all six networks, we first find the best-performing feature layers of each network (with the ridge regression model), and then we compare the results among the networks to select the best network. For each layer of each network, before the ridge regression, we performed PCA and picked the PCA dimension that gave the best results on the validation set.

\section{Results} \label{result}
Surprisingly, we found that features from conv5\textunderscore2 layer of VGG16 trained for object classification slightly outperformed the AlexNet and Inception networks, while the three networks trained solely on faces, VGG-Face, Face-LandmarkNN and Face-SNN did not achieve performance as competitive as the other three for most of the social attributes. The best performing VGG16 layer was conv5\textunderscore2.

We speculate that the reason for the relatively poorer performance of the face recognition networks is that they are optimized either to learn differences between faces which define identity or to learn the face landmark configurations, whereas for this task at hand, we are looking for commonalities behind certain social features which go beyond identity. The landmark network presumably should give results similar to the geometric features, but did not learn features corresponding to all of the features we used in that model. These speculations need to be checked, of course, for example by trying to predict all of our measured features, using the landmark network, but we did not do that here.

We tried fine-tuning the model as follows. We used backpropagation to fine tune the weights into the conv5\textunderscore2 layer, the weights to the PCA layer from conv5\textunderscore2 (initialized by the PCA weights), and the weights from the PCA layer to the output regression unit. However, this fine-tuning did not improve performance, so the results reported in Table \ref{tb:correlations} are without fine-tuning.

We evaluate the performance on 50 random train / validation / test splits of the data with a 64/16/20 percent split for training, cross-validation and testing, respectively. The prediction performance of our model is evaluated using Pearson's correlation with the human ratings on the test set. For each social attribute, we report its human consistency as described in Section \ref{baseline1}. 

Table \ref{tb:correlations} summarizes the prediction performance of our model for all the social attributes compared to Baseline I and II.
The table is organized in a descending order of human agreement on the putative positive attribute of the paired attributes. The three attributes where there is greater agreement among humans for the negative component of the pair are bolded.

Among all the social attributes, human subjects agree most with each other about "happy" and disagree most about "unfamiliar." For both regression models (Baseline II and our model), model performance grows as the consensus on a social trait increases and human correlations with each other are consistently lower than the models' correlations with the average human ratings. Normally, one might consider the human correlations to be an upper bound on performance, but here they are different kinds of correlations. 

Since the change in expression would produce a change in landmark locations, it is not surprising that landmark-based geometric features (Baseline II) achieve comparable or slightly higher correlation as our model for predicting those social attributes that are highly related to expressions (such as "happy", "unhappy", "cold" and "friendly" etc.). While for other social attributes, our model slightly outperforms landmark-based geometric features by about 0.04 correlation on average and significantly outperforms human correlation by about 0.12 correlation on average. This implies that CNN features encode much more information than just landmark-based features. It is essential to visualize those features and understand what features extracted from CNN make our model powerful enough to predict social attributes. 

To quantitatively compare the face social features perceived by humans and those predicted by our best performing model, we take the model predictions on all social features, and compute the Spearman correlation between every pair in the set (see the right figure in Figure \ref{socialCorr}). Not surprisingly, this has very similar patterns compared the heatmap generated from human ratings (see the right panel in Figure \ref{socialCorr}). Pearson Correlation between the upper triangle of the two similarity matrices (human and model prediction) is 0.9836. However, note that each predictor was trained independently.

\begin{center}
\begin{table}[hbt]
{\small
\centering 
\begin{tabular}{|l||c|c|c|}
\hline
\textbf{Social Attributes}& \textbf{Baseline I} & \textbf{Baseline II} & \textbf{Our Model}\\
\hline
happy & 0.84 &\textbf{0.86} & 0.84\\ 
\hline
unhappy & 0.75 & \textbf{0.81} & 0.80\\ 
\hline
friendly & 0.78 &\textbf{0.83} & 0.82\\ 
\hline
unfriendly & 0.72 & \textbf{0.80} & 0.79\\ 
\hline
sociable& 0.74 &\textbf{0.78} & \textbf{0.78}\\ 
\hline
introverted& 0.50 & 0.64 & \textbf{0.65}\\ 
\hline
attractive & 0.72 & 0.66 & \textbf{0.75}\\
\hline
unattractive & 0.62 & 0.62& \textbf{0.70}\\
\hline
kind& 0.72 & \textbf{0.79} & \textbf{0.79}\\ 
\hline
mean& 0.69& \textbf{0.75} & 0.73\\ 
\hline
caring& 0.72 & 0.78 & \textbf{0.79}\\
\hline
cold& 0.71 &\textbf{0.81} & 0.79\\
\hline
trustworthy&0.62 & 0.72 & \textbf{0.73}\\
\hline
untrustworthy&0.60 & 0.69 & \textbf{0.70}\\
\hline
responsible&0.58 &0.65 & \textbf{0.70}\\
\hline
irresponsible&0.55 &0.64& \textbf{0.67}\\
\hline
confident&0.55& 0.55& \textbf{0.61}\\
\hline
uncertain&0.45& 0.62 & \textbf{0.63}\\
\hline
humble&0.55 & \textbf{0.64} & 0.63\\
\hline
egotistic&0.52 & \textbf{0.62} & \textbf{0.62}\\
\hline
emotionally stable&0.53 &0.64 & \textbf{0.67}\\
\hline
emotionally unstable&0.50 &0.62 & \textbf{0.64}\\
\hline
normal &0.49 & 0.58 & \textbf{0.61}\\
\hline
{\bf weird} & 0.52& 0.50 & \textbf{0.56} \\
\hline
intelligent&0.49 & 0.53& \textbf{0.62}\\
\hline
unintelligent&0.43 & 0.53 & \textbf{0.58}\\
\hline
interesting&0.42 & 0.64 & \textbf{0.67}\\
\hline
boring&0.39 & 0.54 & \textbf{0.60}\\
\hline
calm&0.41 & 0.47 & \textbf{0.50}\\ 
\hline
{\bf aggressive}&0.65 &\textbf{0.72} & \textbf{0.72}\\ 
\hline
emotional&0.33 &\textbf{0.60}& \textbf{0.60}\\
\hline
{\bf unemotional}&0.56 &\textbf{ 0.76} & 0.75\\
\hline
memorable&0.30 & 0.38 & \textbf{0.48}\\
\hline
forgettable&0.27 &0.40 &\textbf{ 0.48}\\
\hline
typical&0.28 & 0.41 & \textbf{0.43}\\
\hline
atypical&0.24 & 0.40 & \textbf{0.43}\\
\hline
common &0.25 & 0.37 & \textbf{0.40}\\
\hline
uncommon &0.27 & 0.38 & \textbf{0.40}\\
\hline
familiar&0.24 & 0.42 & \textbf{0.44}\\
\hline
unfamiliar&0.18 & 0.40 & \textbf{0.44}\\
\hline
\end{tabular}}
\caption{Prediction performance of all the social attributes. The reported performance is averaged on 50 random train/validation/test splits of the data.}
\label{tb:correlations}
\end{table}
\end{center}

\section{Feature Visualization} \label{viz}
In this section, we visualize the features that are of importance to social perceptions. We choose facial attractiveness as an example. The same method can be applied to the other social features. We employ the two methods described in section \ref{sec:fv} to visualize features learned by our model. 
\subsection{Data-centric Visualization}
\label{data-centric}
To identify visual features that ignite attractiveness perception, we find the top 9 units of highest influence on attractiveness at conv5\textunderscore2 as follows. First, we compute a product of three terms: (1) A unit's activation from conv5\textunderscore2, (2) that unit's weight to the following fc\textunderscore pca layer, (3) the fc\textunderscore pca unit's weight to the output unit. We then sort all conv5\textunderscore2 units' average products of the three terms and identify the top 9 neurons as the ones that contribute most to the output neuron for the corresponding social feature. 
Then we employ the method described in \cite{yosinski2015understanding,zeiler2014visualizing} to find top-9 input images that cause high activations in each of the top-9 conv5\_2 neurons. Also we further produce the deconvolutional images by projecting each activation separately down to pixel space.

\begin{figure*}[ht]
\centering
\includegraphics[width = \textwidth]{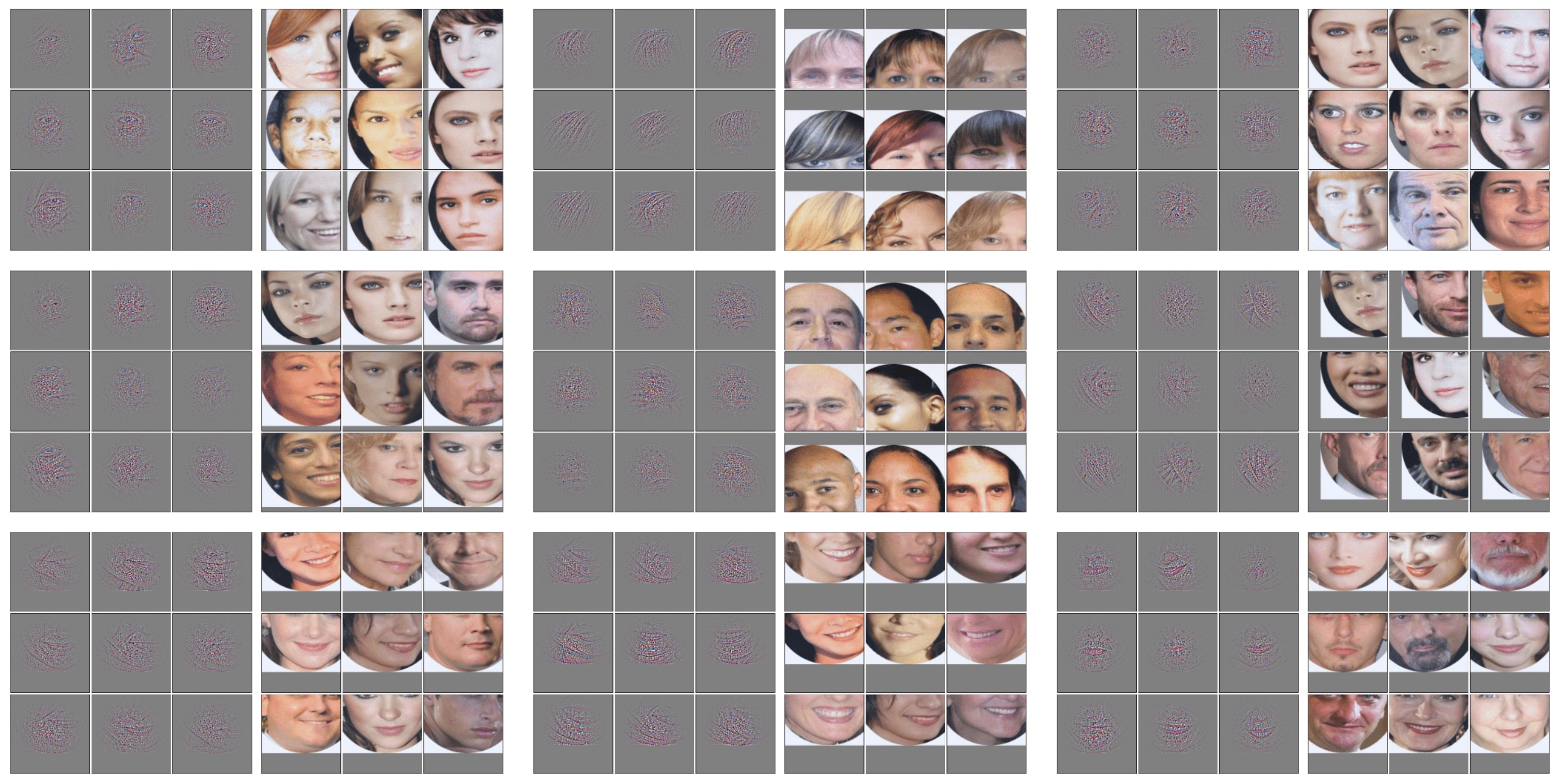}
\caption{Visualization of features in the pretrained-VGG16 regression network. For conv5\textunderscore2 layer, we show the top 9 activations of the top 9 neurons that maximally activate the attractiveness neuron across the training data, projected down to pixel space using the deconvolutional network approach \cite{zeiler2014visualizing} and their corresponding cropped image patches. Best viewed in electronic form, and zoomed in. }
\label{fig:top-9}
\end{figure*}

Figure \ref{fig:top-9} captures the features that are important to predict the attractiveness of a face. The feature importance descends from left to right and  top to bottom. The important features identified by our model are related to eyes, hair with bangs, high nose-bridge, high cheeks, dark eyebrows, strong commanding jawline, chin and red lips. Note that among the 9 cropped input image patches, not all the faces are perceived as attractive or rated as attractive. An attractive face needs to activate more than one feature in order to be considered attractive. This observation agrees with our intuition that attractiveness is a kind of holistic judgment, requiring a combination of multiple features. 

It also seems to be the case that several of the features include relationships between the parts. For example, while the first feature in the upper left of the figure emphasizes the eye, it also includes the nose. This is also true of the upper right feature. Smiling is also important in order to be perceived attractive, as emphasized by the feature in the lower left of the figure.

\subsection{Network-centric Visualization}
In section \ref{data-centric}, we have identified the top-9 units and their feature maps from the con5\_2 layer that maximally activates the attractiveness neuron. Here, we use the gradient-ascent method to optimize the input image that would highly activate a specific neuron of the network. This method is also performed on the pretrained-VGG16 regression model, which is trained to predict attractiveness.

\begin{figure}[!htbp]
\centering
\begin{subfigure}{.3\textwidth}
  \centering
  \includegraphics[width=\linewidth]{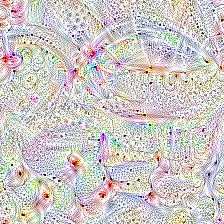}
  \caption{Optimized input image with respect to the output unit}
  \label{fc_out_optimize}
\end{subfigure}
\begin{subfigure}{.5\textwidth}
  \centering
  \includegraphics[width=\linewidth]{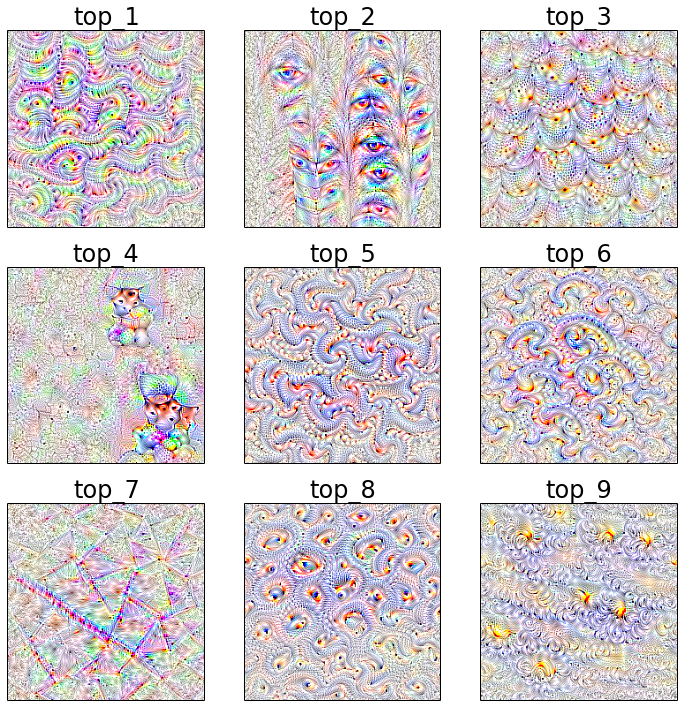}
  \caption{Optimized input images with respect to top-9 neurons from conv5\textunderscore2 layer}
   \label{conv5_2_optimize}
\end{subfigure}
\caption{Visualization of features using network-centric method. To produce Fig \ref{fc_out_optimize}, we use gradient ascent to optimize the output neuron. Fig \ref{conv5_2_optimize} shows 9 optimized images for the feature maps corresponding to top-9 contributor neurons from conv5\textunderscore2 layer.}
\label{network_optimize}
\end{figure}

Figure \ref{fc_out_optimize} shows the optimized image corresponding to the output neuron from a random input image. Optimizing the input image for the output neuron of a regression model does not result in a particularly interpretable figure, although it does appear to emphasize the eyes. Our second approach is to optimize the input image with respect to the top-9 contributing neurons from conv5\textunderscore2 layer that have been identified in section \ref{data-centric}. Figure \ref{conv5_2_optimize} presents 9 optimized images with respect to the corresponding top-9 feature maps of the top-9 neurons from conv5\textunderscore2 layer. Since we use a pretrained-VGG16 network for visualization, it is not surprising that the corresponding top-9 feature maps at conv5\textunderscore2 layer are not particularly encoding facial patterns. 

\begin{figure}[!htbp]
\centering
\begin{subfigure}{.25\textwidth}
  \centering
  \includegraphics[width=.5\linewidth]{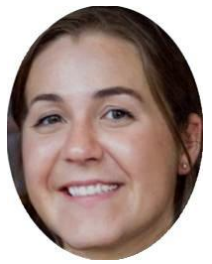}
  \caption{Original input image}
   \label{original}
\end{subfigure}%
\begin{subfigure}{.25\textwidth}
  \centering
  \includegraphics[width=.5\linewidth]{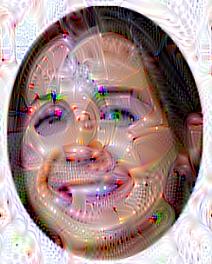} 
   \caption{Optimized image}
  \label{optimize}
\end{subfigure}
\caption{Visualization of the optimized image with a input face image: Figure \ref{original} is the original face image before optimization. Figure \ref{optimize} is produced by performing optimization with respect to the output unit.}
\label{faceOptimize}
\end{figure}

We also present the optimized image initialize with a face image, along with the original face image for comparison in Figure \ref{faceOptimize}. The optimized image tends to highlight the eyes, nose, cheeks and the contour of the face, which is consistent with the features identified by data-centric method.  

\section{Conclusion} \label{conclude}
We have shown that a deep network can be used to predict human social judgments with high correlation with the average human ratings. As far as we know, this is the widest exploration of social judgment predictions, showing human-like perceptions on 40 social dimensions. Unsurprisingly, given previous work recognizing facial expressions, where happiness is the easiest to recognize, our highest correlation is on the happy feature. However, previous work in this area tended to classify a face as happy or not, rather than the degree of rated happiness.

We find that for attributes that correspond to elements of the face that require muscle movement, or a lack of it (such as happy, unhappy, cold, aggressive, unemotional) a simple regression model based on the placement of facial landmarks works well. For ones that don't appear to suggest emotions, such as friendly, note that friendly and happy are highly correlated (see Figure~\ref{socialCorr}, and the red block indexed by happy and friendly). Similarly, aggressive and mean are highly correlated, which presumably requires \emph {not} smiling. 

Perhaps of more significance are the correlations with judgments of traits, such as trustworthiness, responsibleness, confidence, and intelligence,  which would correspond to  more static features of the face. In this area, the deep network, which responds to facial textures  as well as shape, has superior performance. While these judgments do not correspond to a notion of "ground truth," they are things for which humans have a fair amount of agreement, suggesting that there is a signal to be recognized.

Of further note is that we have shown, yet again, that a machine can recognize attractiveness, presumably without any hormonal influences. For this dataset, our deep network correlates with human ratings at 0.75. This provides a benchmark for this dataset. 

Finally, it is of note that we can see that some of the traits considered to be "opposite" in this list are not simply the reverse of one another. For example, there is a large difference in human agreements on "sociable" (0.74) versus "introverted" (0.50), suggesting they are not opposites. 

These results are significant for the field of social robotics. While a robot should not judge a human based completely on their appearance, it can be useful knowledge that humans might judge a person to be trustworthy, while the robot can be more objective. Similarly, a robot need not treat an attractive and unattractive person differently, but this knowledge could affect how the robot interacts with the unattractive person, knowing in advance that this person may have had many negative experiences interacting with people.

In this paper, we train each social feature separately, due to their varied consistency and reliability. In the future, it is worth trying to train one single convnet to learn multiple tasks simultaneously and evaluate whether shared representation may further improve the model performance. 

In summary, we have provided the first machine learning system to learn subjective human judgments of a wide spectrum of traits. We found that the more humans agree on such subjective judgments, the more the system could pick up on the features driving those judgments. It will be of interest to investigate further what those features are, beyond the attractiveness features we displayed here. 

One step further from predicting the value in a certain social feature is to move faces on the social manifold and to increase a face's elicited social perceptions in positive ways (e.g. to make a face look more sociable/ trustworthy/ attractive). Although the images generated by our current visualization method are still far away from being photo-realistic, it may be a fruitful area in the future to develop generative models that can achieve this goal.

{\small
\bibliographystyle{ieee}
\bibliography{egbib}
}

\end{document}